\documentclass[runningheads]{llncs}

\usepackage{accv}

\usepackage{accvabbrv}

\usepackage{amsmath}
\usepackage{arydshln}
\usepackage{booktabs}
\usepackage{enumitem}
\usepackage{float}
\usepackage{graphicx}
\usepackage{lineno}

\usepackage[accsupp]{axessibility}  

\usepackage[pagebackref,breaklinks,colorlinks,citecolor=accvblue]{hyperref}

\usepackage{orcidlink}

\begin{document}

\title{CompareBench: A Benchmark for Visual Comparison Reasoning in Vision-Language Models} 


\author{Jie Cai\orcidlink{0000-0001-6221-0319}}

\authorrunning{J.~Cai}

\institute{OPPO AI Center\\
\email{caijie0620@gmail.com}}

\maketitle

\begin{abstract}
  Visual comparison reasoning is a fundamental capability of vision-language models (VLMs), covering judgments of object quantity, geometric dimensions, spatial relations, and temporal order.
Yet existing benchmarks rarely isolate comparison as a reasoning axis, leaving it unclear whether models can reliably perform comparative visual judgments.
We introduce a benchmark suite organized around three top-level resources: TallyBench, a 2,000-image object counting benchmark; OmniCaps, a 716-image caption and tag resource; and CompareBench, a 1,200-QA visual comparison benchmark.
CompareBench contains four sub-benchmarks spanning quantity, geometric, spatial, and temporal comparison, with the temporal component unifying historical scenes, landmarks, and public figures.
Evaluating nine closed-source model routes from Anthropic, Google, and OpenAI on TallyBench and CompareBench reveals strong overall performance but persistent failures in counting, spatial reasoning, geometric comparison, and temporal ordering.
These results show that visual comparison remains a systematic weakness of current VLMs and establish CompareBench as a focused benchmark for multimodal reasoning evaluation.
All data, code, and prompts will be made publicly available upon acceptance.

\end{abstract}

\section{Introduction}
\label{sec:intro}

\begin{figure*}[t]
  \centering
  \includegraphics[width=\textwidth]{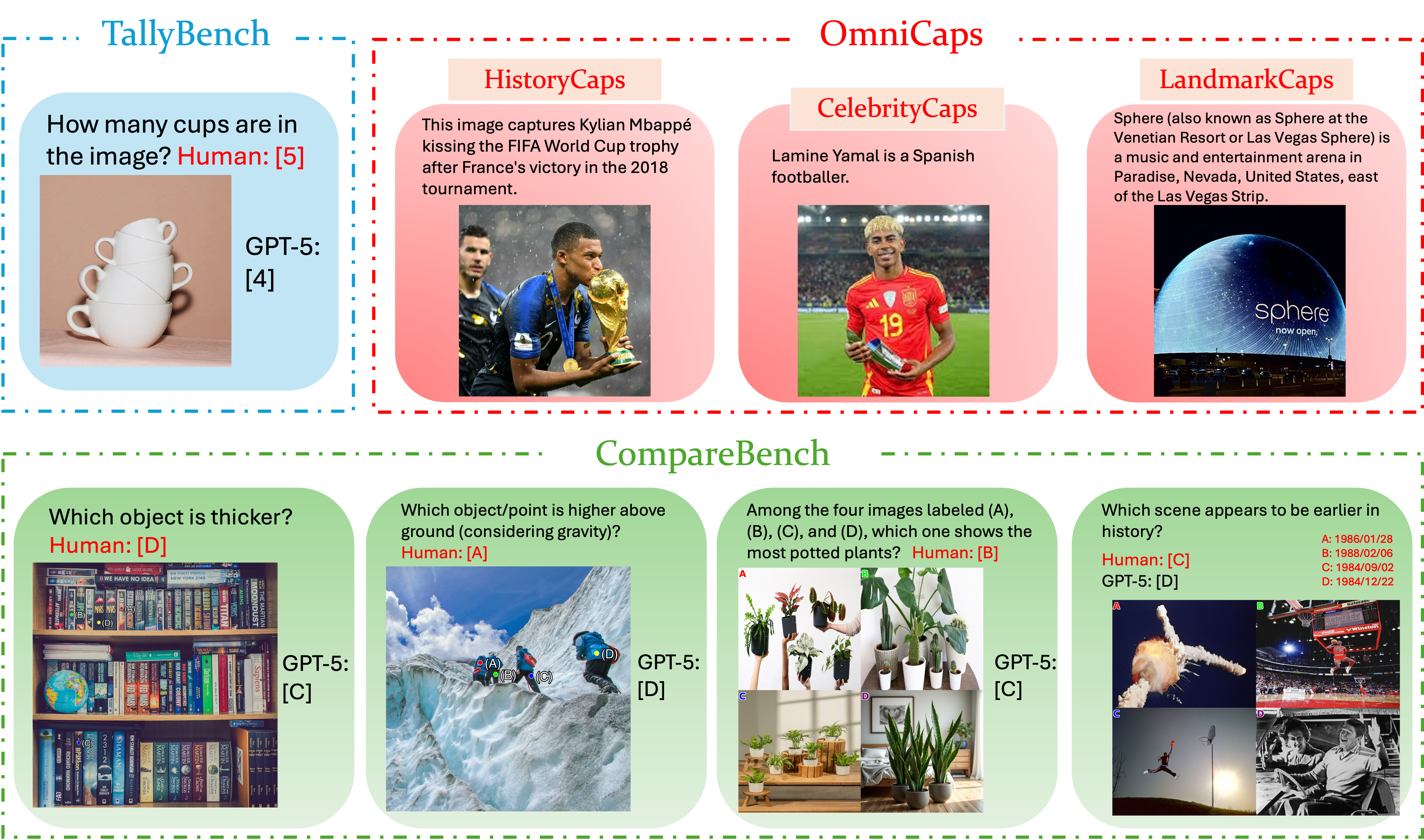}
  \caption{
    Overview of the supporting data and representative GPT-5 failure cases. The top row shows examples from TallyBench and OmniCaps. TallyBench provides object counting data, while OmniCaps includes HistoryCaps, CelebrityCaps, and LandmarkCaps examples with captions. The bottom row shows CompareBench examples covering geometric, spatial, quantity, and temporal comparison. In these cases, GPT-5 gives incorrect answers to questions that are straightforward for humans, including counting stacked cups, comparing book thickness, judging relative height, comparing plant quantities, and ordering historical scenes, which highlights persistent limitations in visual comparison reasoning.
    }
  \label{fig:intro}
\end{figure*}

Vision-language models (VLMs) have achieved strong performance in captioning, visual question answering (VQA), and multimodal reasoning. Visual comparison is a basic capability in human visual understanding, covering judgments about quantity, temporal order, geometric properties, and spatial relations. However, existing VLM evaluations pay limited attention to this capability, leaving comparative reasoning without a systematic assessment.

This gap matters because visual comparison is central to everyday perception, including counting objects, judging depth, and comparing sizes. Existing benchmarks mainly emphasize recognition, description, or commonsense reasoning, while few isolate comparative ability as the target of evaluation. VQA-style datasets focus on open-domain question answering, CLEVR~\cite{johnson2017clevr} emphasizes synthetic logical reasoning, and recent holistic benchmarks such as MMBench~\cite{liu2024mmbench}, MM-Vet~\cite{yu2023mm}, and BLINK~\cite{fu2024blink} test broad multimodal skills, but they do not provide a focused evaluation of comparison reasoning.

As a concrete sign of this gap, Fig.~\ref{fig:intro} shows representative failures of GPT-5 on comparison tasks that are simple for humans. To support targeted evaluation, we organize the data into three top-level resources: CompareBench, TallyBench, and OmniCaps. CompareBench is the main visual comparison QA benchmark and contains four sub-benchmarks: CompareTallyBench, CompareGeometryBench, CompareSpatialBench, and CompareTemporalBench. TallyBench is a 2,000-image object counting benchmark that we evaluate directly with counting accuracy, and it also supports the quantity comparison task in CompareBench. OmniCaps is a 716-image caption and tag resource with three subsets, HistoryCaps, LandmarkCaps, and CelebrityCaps, that support the temporal comparison split and example visualization. Table~\ref{table1} summarizes the CompareBench sub-benchmarks, their task types, scales, formats, and representative questions.

Our contributions are summarized as follows:
\begin{itemize}
\item We organize the data into three top-level resources: CompareBench, TallyBench, and OmniCaps, where CompareBench contains four visual comparison sub-benchmarks, TallyBench serves as a standalone counting benchmark, and OmniCaps contains three caption subsets.
\item We construct CompareBench as the main QA benchmark, forming 1,200 QA pairs that evaluate quantity, temporal, geometric, and spatial comparison.
\item We evaluate closed-source VLMs from OpenAI, Google, and Anthropic on both TallyBench and CompareBench, reporting counting accuracy and comparison accuracy to reveal persistent failures in visual reasoning.
\end{itemize}

\begin{table*}[hbt!]
\centering
\caption{Overview of the four CompareBench sub-benchmarks. The table reports the target reasoning task, sample scale, input format, and a representative question for each subset, covering quantity, geometric, spatial, and temporal comparison. For single-image tasks, each image contains A--D labels marking the candidate objects or points to be compared.}
\label{table1}
\begingroup
\scriptsize
\setlength{\tabcolsep}{2pt}
\renewcommand{\arraystretch}{1.25}
\begin{tabular*}{\textwidth}{@{\extracolsep{\fill}}l c c l p{0.34\textwidth}@{}}
\toprule
Subset & Task & Scale & Format & Example Question \\
\midrule
Compare\-TallyBench    & Quantity & 600 & 4-image grid & Which image shows the most dogs? \\
Compare\-GeometryBench & Geometry & 200 & Single image  & Which object is longer in length? \\
Compare\-SpatialBench  & Spatial  & 100 & Single image  & Which object is closer to the camera? \\
Compare\-TemporalBench & Temporal & 300 & 4-image grid & Which image appears earlier in time? \\
\bottomrule
\end{tabular*}
\endgroup
\end{table*}

\section{Related Work}
\label{sec:related_work}

\subsection{Multimodal Benchmarks}
Existing multimodal benchmarks emphasize recognition, description, question answering, or commonsense reasoning, but do not explicitly target comparative abilities. Captioning datasets such as MSCOCO~\cite{lin2014microsoft} and Flickr30k~\cite{young2014image} are designed for descriptive generation, focusing on aligning visual content with natural language sentences. VQA benchmarks (e.g., VQA and GQA~\cite{hudson2019gqa}) extend beyond description to factual and compositional question answering, testing attributes, objects, or relationships within an image. Reasoning datasets such as RealWorldQA~\cite{realworldqa2024} and ScienceQA~\cite{lu2022learn} incorporate external knowledge and multimodal reasoning, but they rarely test direct comparison across multiple visual entities.

\subsection{Counting and Comparative Reasoning Benchmarks}
Counting has been studied extensively, often as a standalone task. Early works addressed counting in natural scenes or specific domains (e.g., drone-based object counting~\cite{hsieh2017drone}), while TallyQA~\cite{acharya2019tallyqa} posed complex counting questions in a VQA setting. More recent benchmarks explore open-world counting~\cite{amini2023open}, dense counting~\cite{ranjan2021learning}, or teaching CLIP to count~\cite{paiss2023teaching}. However, these benchmarks remain limited to numerosity and do not generalize to broader comparative reasoning tasks such as size, distance, or temporal order.

Synthetic datasets such as CLEVR~\cite{johnson2017clevr} provide carefully designed tests for compositional logic and spatial reasoning, but their artificial nature and limited diversity hinder transfer to real-world applications. More holistic benchmarks such as MMBench~\cite{liu2024mmbench}, MM-Vet~\cite{yu2023mm}, BLINK~\cite{fu2024blink}, and SimpleVQA~\cite{cheng2025simplevqa} have recently emerged to test a wide range of multimodal skills, from recognition and perception to knowledge grounding and factuality. Yet, even in these comprehensive evaluations, comparative reasoning is not systematically represented, and questions about ``which object is larger/closer/earlier'' are often absent. In contrast, our benchmark suite is designed to fill this gap through three top-level resources: TallyBench for standalone object counting evaluation, OmniCaps for date-keyed captions and tags, and CompareBench for visual comparison QA. CompareBench provides a controlled and diverse evaluation of four fundamental comparison tasks (quantity, temporal, geometric, and spatial reasoning), making it a complementary diagnostic tool for evaluating a core but underexplored dimension of VLMs.

\section{Methodology}
\label{sec:methodology}

\subsection{OmniCaps (716 Images)}
\label{sec:omnicaps}

OmniCaps is a multi-domain caption and tag resource with 716 date-keyed images. It contains three subsets: HistoryCaps, LandmarkCaps, and CelebrityCaps. Each image is named by its associated date and paired with metadata fields including an image identifier, a short tag, and a caption. HistoryCaps contains 516 historical event images spanning 1707--2025, LandmarkCaps contains 100 world landmark images with construction or opening dates from 80--2024, and CelebrityCaps contains 100 notable people images with birth dates from 1643--2007.

OmniCaps supports the construction of CompareTemporalBench. HistoryCaps provides historical scene and event examples, LandmarkCaps provides landmark images with construction or opening dates, and CelebrityCaps provides public-figure images with birth dates. These three sources are merged into one temporal split, where date-keyed metadata provides the ground-truth temporal order and captions provide contextual descriptions of the image content.

\subsection{TallyBench (2,000 Samples)}
\label{sec:tallybench}

\begin{figure*}[hbt!]
  \centering
  \includegraphics[width=1.0\textwidth]{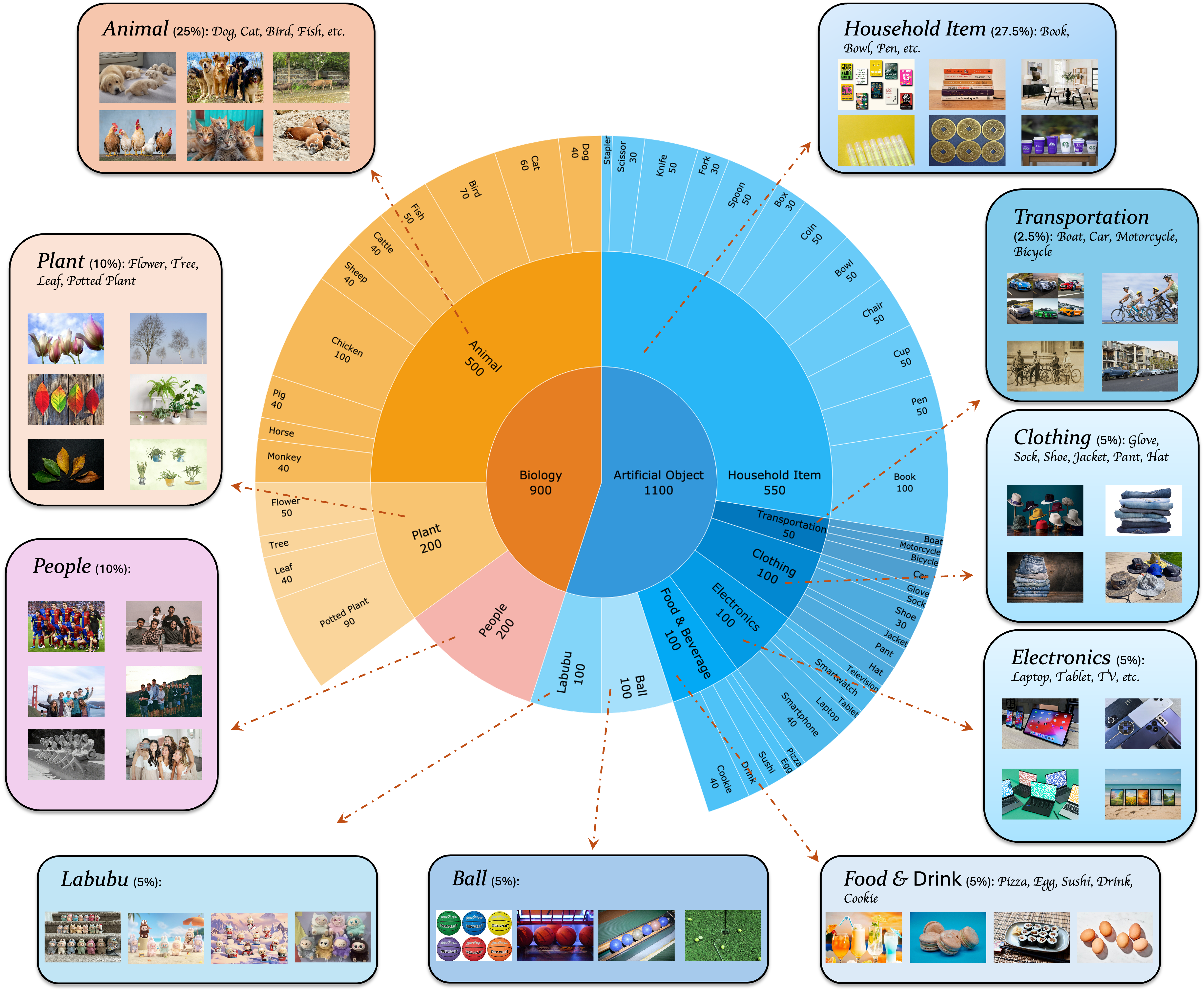}
  \caption{Distribution of TallyBench categories. The top level separates biological objects from artificial objects, while the outer level shows fine-grained categories such as dogs, cats, chickens, books, spoons, knives, plants, balls, and household items.}
  \label{fig:tallybench-dist}
\end{figure*}

TallyBench is a standalone object counting benchmark with 2,000 images, each paired with a counting question and an integer ground-truth answer. Each sample is stored as a JSON entry with \texttt{image\_name}, \texttt{vlm\_question}, \texttt{gt\_answer}, \texttt{categories}, and \texttt{image\_type}. As shown in Fig.~\ref{fig:tallybench-dist}, TallyBench covers 49 fine-grained categories across biological and artificial objects, including animals, plants, people, food and beverages, electronics, clothing, transportation, balls, household items, and other common categories.

To ensure consistent evaluation, each image-question pair is paired with a unified instruction template. The instruction requires models to count all clearly identifiable instances of the target category, include partially occluded or cropped objects when they remain recognizable, exclude reflections, and output only a single integer. These rules make TallyBench an independent benchmark for counting accuracy and also provide the source images for the quantity comparison task in CompareBench.

\subsection{CompareBench (1,200 Samples)}
\label{sec:comparebench}

\begin{figure*}[hbt!]
  \centering
  \includegraphics[width=1.0\textwidth]{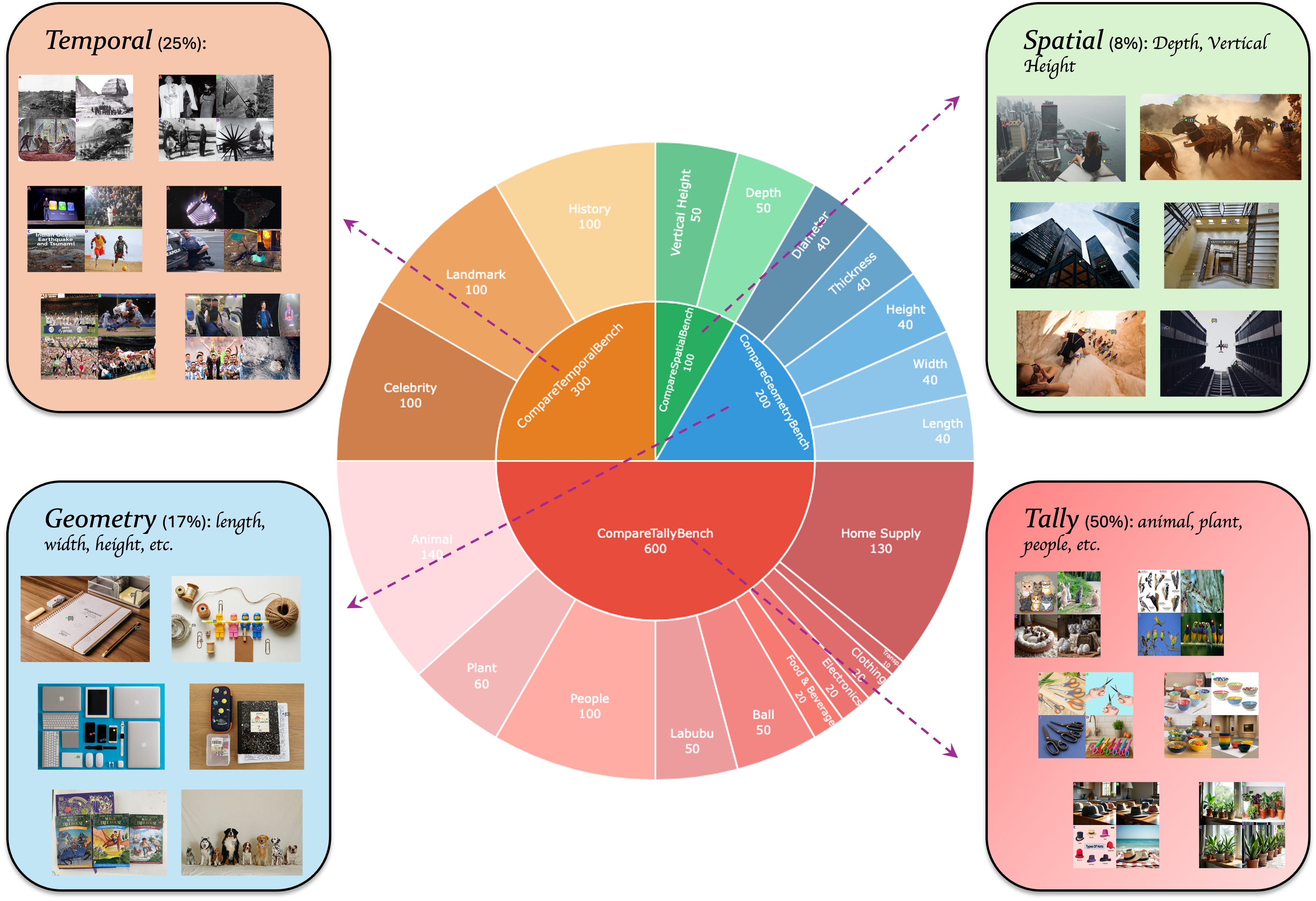}
  \caption{Category distribution of CompareBench. The benchmark covers quantity, geometric, spatial, and temporal comparison. Quantity samples are derived from TallyBench, geometric samples cover length, width, height, thickness, and diameter, spatial samples cover depth and vertical height, and temporal samples are derived from OmniCaps subsets.}
  \label{fig:comparebench-dist}
\end{figure*}

CompareBench is the main visual comparison QA benchmark. It contains 1,200 samples across four sub-benchmarks and covers four comparison abilities: quantity, geometric, spatial, and temporal reasoning. CompareTallyBench is built from TallyBench, CompareTemporalBench is built from OmniCaps subsets, and CompareGeometryBench and CompareSpatialBench are built from additional annotated images. Each task uses a standardized instruction template from the prompt configuration, and models are required to output only a single choice (A--D).

\textbf{CompareGeometryBench (200)} targets geometric comparison. Each sample consists of a single image with four candidate objects or points labeled A--D. The questions ask models to compare geometric properties such as length, width, height, thickness, or diameter. Models must consider only the labeled candidates and answer with a single letter.

\textbf{CompareSpatialBench (100)} evaluates spatial relation reasoning. Each sample is a single image with four labeled points or objects. The questions focus on relative depth or vertical height, such as which object is closer to the camera or which point is higher above the ground. Models must compare only the labeled locations and ignore unrelated image context.

\textbf{CompareTallyBench (600)} targets quantity comparison. Each sample is a $1600 \times 1600$ four-image grid constructed from TallyBench images. The model must compare object counts across the four sub-images and identify the option with the requested maximum or minimum count.

\textbf{CompareTemporalBench (300)} targets temporal comparison. It merges three 100-sample temporal scenarios from OmniCaps: historical scenes and events from HistoryCaps, landmarks ordered by construction or opening date from LandmarkCaps, and public figures ordered by birth date from CelebrityCaps. The merged split uses image names 0001--0100 for historical scenes, 0101--0200 for landmarks, and 0201--0300 for public figures.

\textbf{Together}, the four sub-benchmarks provide a systematic framework for evaluating visual comparison in VLMs. By covering quantity, temporal, geometric, and spatial reasoning, CompareBench targets a core but underexplored capability and complements recognition- and description-oriented benchmarks.

\section{Experiments}
\label{sec:experiments}

\subsection{Results and Analysis}

We evaluated representative closed-source VLMs from OpenAI, Gemini, and Claude. Together, these models cover the strongest widely used API-based systems for multimodal reasoning.  

\textbf{Results.} Accuracy (\%) is used as the unified evaluation metric across all benchmarks.  
For TallyBench, accuracy is defined as the percentage of exact integer matches between predicted and ground-truth counts.  
For CompareBench, each question is formulated as a single-choice (A, B, C, or D) with exactly one correct answer, and accuracy is measured as the proportion of correct predictions.  

\begin{table}[hbt!]
\centering
\caption{Accuracy (\%) on TallyBench. 
TallyBench contains 2,000 single-image counting questions, and accuracy is computed by exact integer match between the predicted count and the ground-truth count. 
We report representative closed-source VLMs from Claude, OpenAI, and Gemini, together with human performance.}
\label{table2}
\begin{tabular}{lcc}
\toprule
Model & TallyBench (2,000) \\
\midrule
Claude Haiku 4.5       & 54.55  \\
Claude Sonnet 4.6      & 68.00  \\
Claude Opus 4.8        & 75.40  \\
\addlinespace[6pt]
OpenAI GPT-5.4-mini    & 83.90  \\
OpenAI GPT-5.4         & 86.85  \\
OpenAI GPT-5.5         & 87.15  \\
\addlinespace[6pt]
Gemini 3.1 Flash-Lite  & 79.60  \\
Gemini 3.5 Flash       & 96.15  \\
Gemini 3.1 Pro         & 94.60  \\
\midrule
Human    & 98.00   \\
\bottomrule
\end{tabular}
\end{table}

\begin{table*}[hbt!]
\centering
\caption{Accuracy (\%) on CompareBench. 
CompareBench contains 1,200 multiple-choice comparison questions across four sub-benchmarks: Tally = CompareTallyBench, Geometry = CompareGeometryBench, Spatial = CompareSpatialBench, and Temporal = CompareTemporalBench. 
Numbers in parentheses indicate the number of samples in each sub-benchmark.}
\label{table3}
\begin{tabular}{lccccc}
\toprule
Model & \shortstack{Tally\\(600)} & \shortstack{Geometry\\(200)} & \shortstack{Spatial\\(100)} & \shortstack{Temporal\\(300)} & \shortstack{Average\\(1,200)} \\
\midrule
Claude Haiku 4.5     & 46.17 & 39.50 & 34.00 & 54.00 & 46.00 \\
Claude Sonnet 4.6    & 65.00 & 55.00 & 62.00 & 67.00 & 63.58 \\
Claude Opus 4.8      & 75.83 & 69.50 & 75.00 & 66.67 & 72.42 \\
\addlinespace[6pt]
OpenAI GPT-5.4-mini  & 85.17 & 79.00 & 81.00 & 80.33 & 82.58 \\
OpenAI GPT-5.4       & 88.67 & 86.50 & 85.00 & 84.67 & 87.00 \\
OpenAI GPT-5.5       & 86.33 & 82.00 & 85.00 & 79.00 & 83.67 \\
\addlinespace[6pt]
Gemini 3.1 Flash-Lite & 78.67 & 72.00 & 82.00 & 76.67 & 77.33 \\
Gemini 3.5 Flash      & 97.00 & 87.50 & 93.00 & 89.33 & 93.17 \\
Gemini 3.1 Pro        & 96.17 & 84.00 & 88.00 & 88.67 & 91.58 \\
\midrule
Human & 99.00 & 98.00 & 98.00 & 31.67 & 81.92 \\
\bottomrule
\end{tabular}
\end{table*}

Table~\ref{table2} and Table~\ref{table3} report the performance of all evaluated models on TallyBench and the four CompareBench sub-benchmarks: CompareTallyBench, CompareGeometryBench, CompareSpatialBench, and CompareTemporalBench.
On TallyBench, human annotators achieve the highest accuracy, while Gemini~3.5 Flash and Gemini~3.1 Pro are the strongest model baselines.
On CompareBench, Gemini~3.5 Flash obtains the best average accuracy, followed by Gemini~3.1 Pro and OpenAI GPT-5.4.
Although humans remain near ceiling on counting, geometric, and spatial comparisons, their accuracy drops sharply on CompareTemporalBench, indicating that temporal comparison often requires external world knowledge beyond direct visual comparison.

\begin{figure*}[hbt!]
\centering
\includegraphics[width=\textwidth]{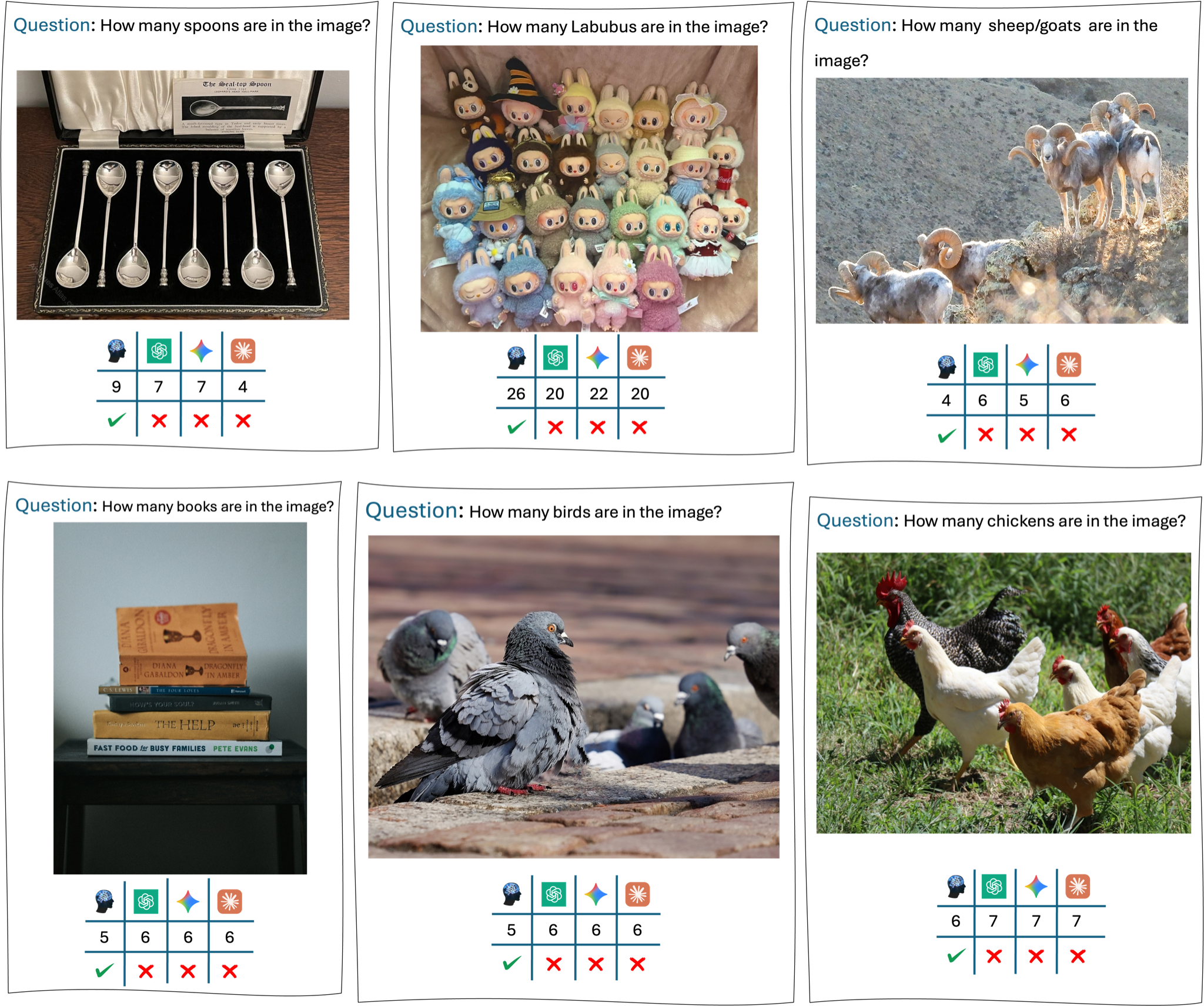}
\caption{\textbf{TallyBench hard cases where all three models fail.}
Each panel shows the image, the counting question (e.g., ``How many spoons are in the image?''), and the predictions from three models, all of which are incorrect.
The six examples (top-left to bottom-right) cover \emph{spoons}, \emph{Labubu} instances, \emph{sheep/goats}, \emph{stacked books}, \emph{birds}, and \emph{chickens}.
These cases illustrate typical counting failure modes, including confusing visually similar instances, missing partially occluded objects, and misreading fine-scale duplicates, despite such tasks being trivial for humans.}
\label{fig:tallybench-hardcases}
\end{figure*}

\begin{figure*}[hbt!]
\centering
\includegraphics[width=\textwidth]{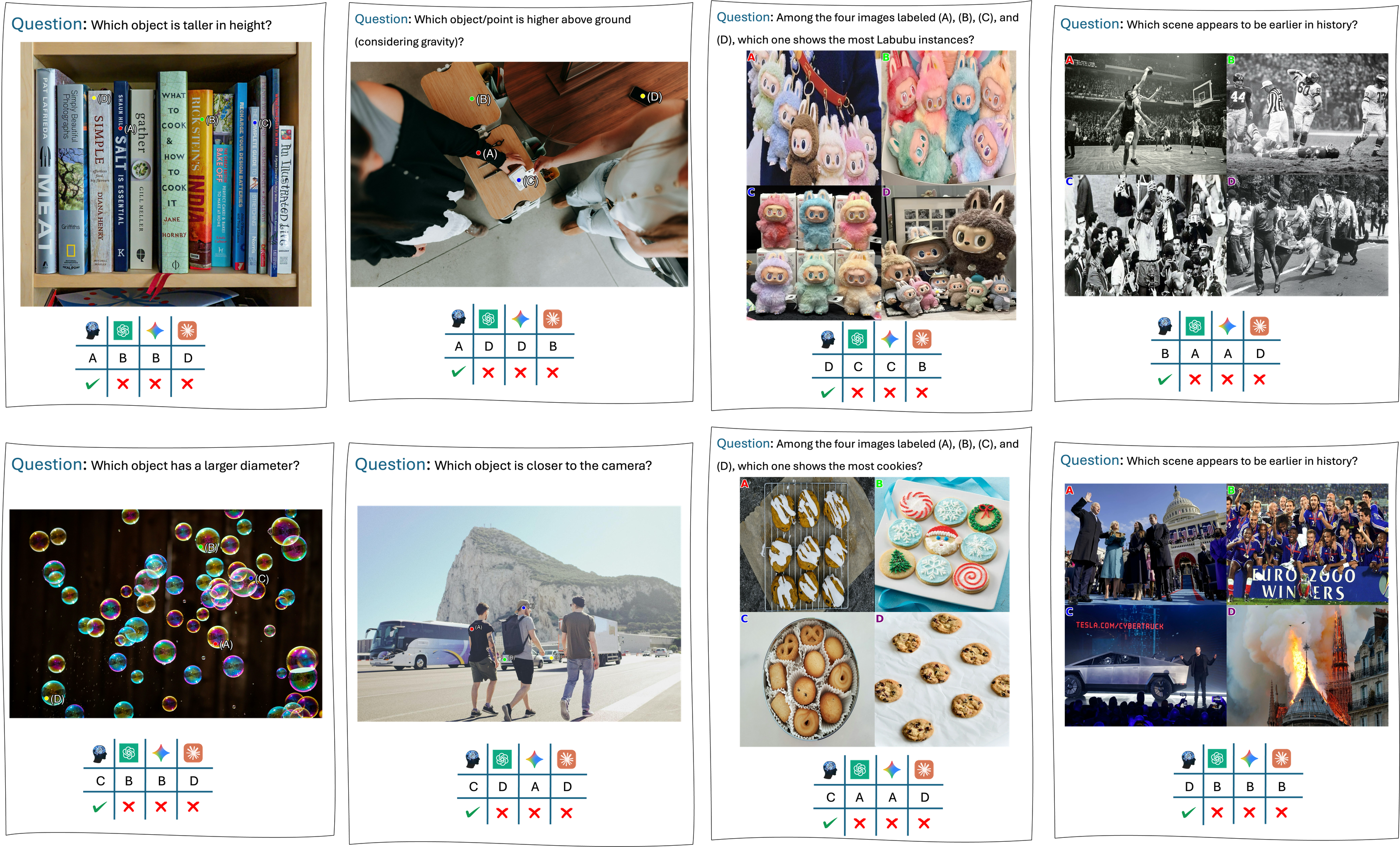}
\caption{\textbf{Failure cases from CompareBench.} 
Each panel shows a sample question and predictions from three state-of-the-art VLMs, all of which are incorrect. 
From left to right on the first line: (\emph{Geometry}) identifying the taller book; 
(\emph{Spatial}) deciding which marked point is higher above ground; 
(\emph{Tally}) comparing the quantity of Labubus; 
(\emph{Temporal}) selecting the earliest historical scene. 
These cases highlight persistent weaknesses in comparative reasoning across dimensions of size, space, quantity, and time.}
\label{fig:comparebench-hardcases}
\end{figure*}

\textbf{Analysis.} As shown in Fig.~\ref{fig:tallybench-hardcases}, even the strongest VLMs simultaneously fail on simple TallyBench cases such as spoons, books, birds, and chickens, which are trivial for humans. Similarly, Fig.~\ref{fig:comparebench-hardcases} illustrates typical failure cases from CompareBench across geometric, spatial, quantity, and temporal reasoning. Despite strong overall performance, all three strong VLMs misinterpret tasks that require distinguishing object length versus height, or fail to classify foreground and background correctly in spatial reasoning.  
Counting-related tasks (TallyBench and CompareTallyBench) are relatively easier, while temporal and spatial reasoning remain the most challenging. These results indicate that current VLMs still lack robust comparative reasoning capabilities: they struggle not only on tasks trivial for humans but also, in certain cases such as temporal reasoning, succeed only by leveraging background knowledge that humans may not apply without external context.  

A striking observation from CompareBench is the discrepancy between human and model performance on temporal reasoning.  
While humans achieve near-perfect accuracy on counting, geometric, and spatial comparisons, their performance drops sharply on CompareTemporalBench. This is because many temporal samples involve visually similar scenes, identities, or landmarks where chronological ordering requires historical knowledge beyond the image itself (e.g., subtle differences in clothing, architecture, technology, or construction dates). In contrast, large-scale VLMs such as Gemini~3.5 Flash and Gemini~3.1 Pro achieve substantially higher scores, likely by leveraging background knowledge learned from massive pretraining corpora. This divergence highlights that CompareTemporalBench does not merely test perceptual comparison but also evaluates the integration of visual evidence with external world knowledge.  
Consequently, it provides a unique diagnostic setting where VLMs can, in some cases, surpass human annotators constrained to vision-only reasoning.

\subsection{Model Performance and Cost}
\label{sec:cost_performance}

\begin{figure*}[hbt!]
\centering
\includegraphics[width=\textwidth]{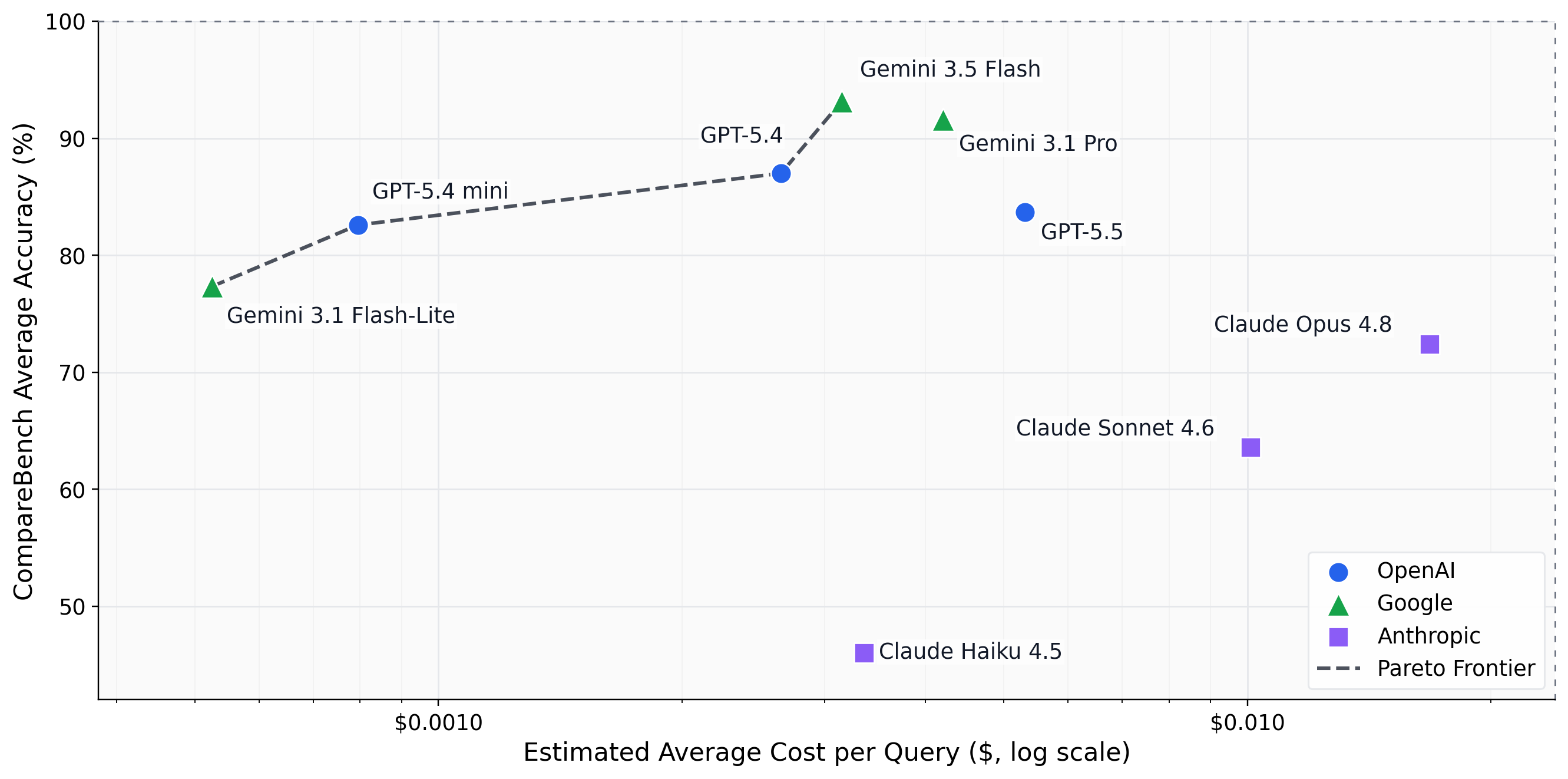}
\caption{\textbf{Cost-performance tradeoff on CompareBench.}
The x-axis reports the estimated Average Cost per Query in USD, including image input, text input, and a one-token text output, and the y-axis reports CompareBench Average Accuracy.
The dashed line shows the Pareto Frontier across the evaluated model routes.}
\label{fig:cost_performance}
\end{figure*}

We estimate the average per-query inference cost on CompareBench, including image input, text input, and the short text output, while excluding retries, caching, and provider-specific overhead. Across the nine evaluated model routes, the mean estimated cost is \$0.00521 per query, and Figure~\ref{fig:cost_performance} shows that Gemini~3.5 Flash provides the strongest cost-performance point, achieving 93.17\% average accuracy at approximately \$0.00315 per query. The provider-level averages are \$0.00263 for Google, \$0.00292 for OpenAI, and \$0.01008 for Anthropic, while the Pareto Frontier consists of Gemini~3.1 Flash-Lite, GPT-5.4 mini, GPT-5.4, and Gemini~3.5 Flash. These results indicate that token price and benchmark accuracy are not monotonic under the fixed CompareBench prompt protocol.

\section{Conclusion}
\label{sec:conclusion}

We introduced CompareBench, a benchmark of 1,200 QA pairs across four sub-benchmarks targeting four fundamental visual comparison tasks: quantity, temporal, geometric, and spatial reasoning, built from TallyBench (2,000 counting images) and OmniCaps (716 date-keyed images covering historical events, landmarks, and celebrities). 
Experiments with representative closed-source VLMs from OpenAI, Gemini, and Claude show that strong API-based systems still make systematic errors on fine-grained comparison tasks, including object counting, geometric comparison, and spatial reasoning. 
A notable finding is the knowledge-driven reversal in CompareTemporalBench, where VLMs can surpass human accuracy by drawing on encyclopedic pretraining knowledge that individual human annotators rarely possess in full. 
CompareBench thus provides a principled testbed for diagnosing both perception-driven limitations (e.g., confusing object length with height, missing small counting differences, or misclassifying foreground and background) and knowledge-driven challenges (e.g., ordering historical events, celebrities, or landmarks), making it a targeted complement to existing multimodal benchmarks.
We hope future work will build on CompareBench to incorporate fine-grained difficulty and error-type annotations for more targeted failure analysis, and to extend its comparison dimensions beyond the current four to include attributes such as color, texture, and material properties.

\section*{Limitations}
\label{sec:limitations}

\textbf{Benchmark scale and coverage.}
CompareBench currently contains 1,200 QA pairs covering four comparison tasks: quantity, geometry, spatial reasoning, and temporal ordering, with TallyBench and OmniCaps providing the supporting data for counting and temporal comparison.
The benchmark emphasizes manually constructed samples and verifiable answers, so it does not cover all possible comparison dimensions, visual domains, linguistic formulations, or real-world application distributions.
Therefore, the results should be interpreted as a focused diagnostic evaluation of current VLM comparison reasoning rather than a complete assessment of all visual comparison abilities.

\textbf{Temporal comparison requires external knowledge.}
Samples in CompareTemporalBench often require information beyond direct visual evidence, including historical dates, landmark construction or opening years, and the birth years of public figures.
We keep this design intentionally because temporal ordering in real multimodal settings often requires models to combine visual evidence with background knowledge.
As a result, the temporal split should not be viewed as a pure test of visual temporal perception, and this knowledge dependence also explains why VLMs can outperform a human baseline that relies mainly on visual reasoning.

\textbf{Model and evaluation coverage.}
Our experiments evaluate nine closed-source API model routes from Anthropic, Google, and OpenAI under a fixed prompt protocol, using accuracy as the primary metric.
Open-source VLMs, task-specific fine-tuned models, multi-turn interaction settings, visible reasoning traces, and prompt sensitivity are not systematically included in the current evaluation.
The reported findings therefore mainly characterize representative closed-source VLMs under this evaluation protocol, and future work should expand the model families, evaluation settings, and fine-grained error annotations to test the generality of these conclusions.

%
%
\bibliographystyle{splncs04}
\bibliography{main}

@String(BMVC  = {Brit. Mach. Vis. Conf.})

@String(AAAI  = {AAAI})

@String(BMVC  =	{BMVC})

@inproceedings{lin2014microsoft,
  title={Microsoft coco: Common objects in context},
  author={Lin, Tsung-Yi and Maire, Michael and Belongie, Serge and Hays, James and Perona, Pietro and Ramanan, Deva and Doll{\'a}r, Piotr and Zitnick, C Lawrence},
  booktitle={European conference on computer vision},
  pages={740--755},
  year={2014},
  organization={Springer}
}

@misc{realworldqa2024,
  title        = {RealWorldQA: A Benchmark for Real-World Spatial Understanding in Vision-Language Models},
  author       = {xAI},

  year         = {2024},
  howpublished = {\url{https://huggingface.co/datasets/visheratin/realworldqa}},
  note         = {Accessed: 2025-08-25}
}

@article{cheng2025simplevqa,
  title={Simplevqa: Multimodal factuality evaluation for multimodal large language models},
  author={Cheng, Xianfu and Zhang, Wei and Zhang, Shiwei and Yang, Jian and Guan, Xiangyuan and Wu, Xianjie and Li, Xiang and Zhang, Ge and Liu, Jiaheng and Mai, Yuying and others},
  journal={arXiv preprint arXiv:2502.13059},
  year={2025}
}

@article{young2014image,
  title={From image descriptions to visual denotations: New similarity metrics for semantic inference over event descriptions},
  author={Young, Peter and Lai, Alice and Hodosh, Micah and Hockenmaier, Julia},
  journal={Transactions of the association for computational linguistics},
  volume={2},
  pages={67--78},
  year={2014},
  publisher={MIT Press One Rogers Street, Cambridge, MA 02142-1209, USA journals-info~…}
}

@article{lu2022learn,
  title={Learn to explain: Multimodal reasoning via thought chains for science question answering},
  author={Lu, Pan and Mishra, Swaroop and Xia, Tanglin and Qiu, Liang and Chang, Kai-Wei and Zhu, Song-Chun and Tafjord, Oyvind and Clark, Peter and Kalyan, Ashwin},
  journal={Advances in Neural Information Processing Systems},
  volume={35},
  pages={2507--2521},
  year={2022}
}

@inproceedings{liu2024mmbench,
  title={Mmbench: Is your multi-modal model an all-around player?},
  author={Liu, Yuan and Duan, Haodong and Zhang, Yuanhan and Li, Bo and Zhang, Songyang and Zhao, Wangbo and Yuan, Yike and Wang, Jiaqi and He, Conghui and Liu, Ziwei and others},
  booktitle={European conference on computer vision},
  pages={216--233},
  year={2024},
  organization={Springer}
}

@article{yu2023mm,
  title={Mm-vet: Evaluating large multimodal models for integrated capabilities},
  author={Yu, Weihao and Yang, Zhengyuan and Li, Linjie and Wang, Jianfeng and Lin, Kevin and Liu, Zicheng and Wang, Xinchao and Wang, Lijuan},
  journal={arXiv preprint arXiv:2308.02490},
  year={2023}
}

@inproceedings{acharya2019tallyqa,
  title={Tallyqa: Answering complex counting questions},
  author={Acharya, Manoj and Kafle, Kushal and Kanan, Christopher},
  booktitle={Proceedings of the AAAI conference on artificial intelligence},
  volume={33},
  pages={8076--8084},
  year={2019}
}

@article{amini2023open,
  title={Open-world Text-specified Object Counting},
  author={Amini-Naieni, Niki and Amini-Naieni, Kiana and Han, Tengda and Zisserman, Andrew},
  journal={BMVC},
  year={2023}
}

@inproceedings{hudson2019gqa,
  title={Gqa: A new dataset for real-world visual reasoning and compositional question answering},
  author={Hudson, Drew A and Manning, Christopher D},
  booktitle={Proceedings of the IEEE/CVF conference on computer vision and pattern recognition},
  pages={6700--6709},
  year={2019}
}

@inproceedings{ranjan2021learning,
  title={Learning to count everything},
  author={Ranjan, Viresh and Sharma, Udbhav and Nguyen, Thu and Hoai, Minh},
  booktitle={Proceedings of the IEEE/CVF Conference on Computer Vision and Pattern Recognition},
  pages={3394--3403},
  year={2021}
}

@inproceedings{fu2024blink,
  title={Blink: Multimodal large language models can see but not perceive},
  author={Fu, Xingyu and Hu, Yushi and Li, Bangzheng and Feng, Yu and Wang, Haoyu and Lin, Xudong and Roth, Dan and Smith, Noah A and Ma, Wei-Chiu and Krishna, Ranjay},
  booktitle={European Conference on Computer Vision},
  pages={148--166},
  year={2024},
  organization={Springer}
}

@inproceedings{hsieh2017drone,
  title={Drone-based object counting by spatially regularized regional proposal network},
  author={Hsieh, Meng-Ru and Lin, Yen-Liang and Hsu, Winston H},
  booktitle={Proceedings of the IEEE international conference on computer vision},
  pages={4145--4153},
  year={2017}
}

@inproceedings{paiss2023teaching,
  title={Teaching clip to count to ten},
  author={Paiss, Roni and Ephrat, Ariel and Tov, Omer and Zada, Shiran and Mosseri, Inbar and Irani, Michal and Dekel, Tali},
  booktitle={Proceedings of the IEEE/CVF International Conference on Computer Vision},
  pages={3170--3180},
  year={2023}
}

@inproceedings{johnson2017clevr,
  title={Clevr: A diagnostic dataset for compositional language and elementary visual reasoning},
  author={Johnson, Justin and Hariharan, Bharath and Van Der Maaten, Laurens and Fei-Fei, Li and Lawrence Zitnick, C and Girshick, Ross},
  booktitle={Proceedings of the IEEE conference on computer vision and pattern recognition},
  pages={2901--2910},
  year={2017}
}
\end{document}